\documentclass[letterpaper, 10 pt, conference]{ieeeconf}
\IEEEoverridecommandlockouts

\usepackage{cite}
\usepackage{amsmath,amssymb,amsfonts}
\usepackage{algorithmic}
\usepackage{graphicx}
\usepackage{textcomp}
\usepackage[dvipsnames]{xcolor}
\usepackage[font=footnotesize]{caption}
\usepackage{subcaption}
\usepackage{booktabs}
\usepackage{siunitx}
\usepackage{makecell}
\usepackage{hyperref}
\hypersetup{
    colorlinks=true,
    linkcolor=blue!80!black,
    filecolor=blue!80!black,
    urlcolor=blue!80!black,
    citecolor=blue!80!black}

\newbool{DRAFT}
\setbool{DRAFT}{true}
\newcommand{\COMMENT}[3]{\ifbool{DRAFT}{\textcolor{#1}{[#2 -#3]}}{}}

\def\BibTeX{{\rm B\kern-.05em{\sc i\kern-.025em b}\kern-.08em
    T\kern-.1667em\lower.7ex\hbox{E}\kern-.125emX}}
\begin{document}

\title{Hearing the Slide: Acoustic-Guided Constraint Learning for Fast Non-Prehensile Transport}

\author{ Yuemin Mao$^{1}$, Bardienus P. Duisterhof$^{1}$, Moonyoung Lee$^{1}$,
 Jeffrey Ichnowski$^{1}$
\thanks{$^{1}$ Robotics Institute, Carnegie Mellon University. {\tt\small \{yueminm, jeffi\}@andrew.cmu.edu}}%
}

\maketitle


\begin{abstract}
Object transport tasks are fundamental in robotic automation, emphasizing the importance of efficient and secure methods for moving objects. Non-prehensile transport can significantly improve transport efficiency, as it enables handling multiple objects simultaneously and accommodating objects unsuitable for parallel-jaw or suction grasps. Existing approaches incorporate constraints based on the Coulomb friction model, which is imprecise during fast motions where inherent mechanical vibrations occur. Imprecise constraints can cause transported objects to slide or even fall off the tray. To address this limitation, we propose a novel method to learn a friction model using acoustic sensing that maps a tray's motion profile to a dynamically conditioned friction coefficient. This learned model enables an optimization-based motion planner to adjust the friction constraint at each control step according to the planned motion at that step. In experiments, we generate time-optimized trajectories for a UR5e robot to transport various objects with constraints using both the standard Coulomb friction model and the learned friction model. Results suggest that the learned friction model reduces object displacement by up to 86.0\,\% compared to the baseline, highlighting the effectiveness of acoustic sensing in learning real-world friction constraints. Video demonstrations see \href{https://fast-non-prehensile.github.io/}{https://fast-non-prehensile.github.io/}. 
\end{abstract}

\section{Introduction}
When using a tray as the end effector, robots avoid issues around parallel jaw grippers where objects can be too large for the gripper~\cite{mahler2016dex}, difficult-to-grasp~\cite{wang2019adversarial}, or irregularly-shaped~\cite{morrison2020egadevolvedgraspinganalysis}. Similarly, trays avoid issues around suction grasping porous or deformable objects, or computing suction-grasp stability~\cite{mahler2018dex, DBLP:conf/wafr/AvigalICG22}. Additionally, non-prehensile transport can boost efficiency by allowing simultaneous transport of multiple objects~\cite{agboh2022multiobjectgraspingplane, agboh2023learningefficientlyplanrobust}, reduce computational costs by avoiding computations for traditional rigid grasps, and lower upfront and maintenance expenses as trays have no actuation. 

We consider a modular workcell setup where a robotic manipulator uses a tray to transport objects, adaptable to a wide range of automated manufacturing scenarios that benefit from fast object handling. A key safety concern, especially when moving large or fragile objects, is minimizing any displacement of the object on the tray, which requires balancing inertial and frictional forces on the transported objects. Recent works have addressed the challenges of fast non-prehensile transport by incorporating constraints based on the Coulomb friction model with model predictive control (MPC) to secure objects~\cite{SelvaggioTCST2023, Heins_2023}. However, they do not explicitly aim to push the robots to their speed limit or minimize object sliding, which occurs when the Coulomb friction model fails to account for some real-world factors during fast transport. We address this limitation by augmenting the Coulomb friction model. We analyze the dynamic interactions between the tray and transported objects to develop a friction model that infers the real-world friction constraint, allowing a robot to simultaneously minimize object sliding and transport time.

\begin{figure}[t]
    \centering
    \includegraphics[width=\linewidth]{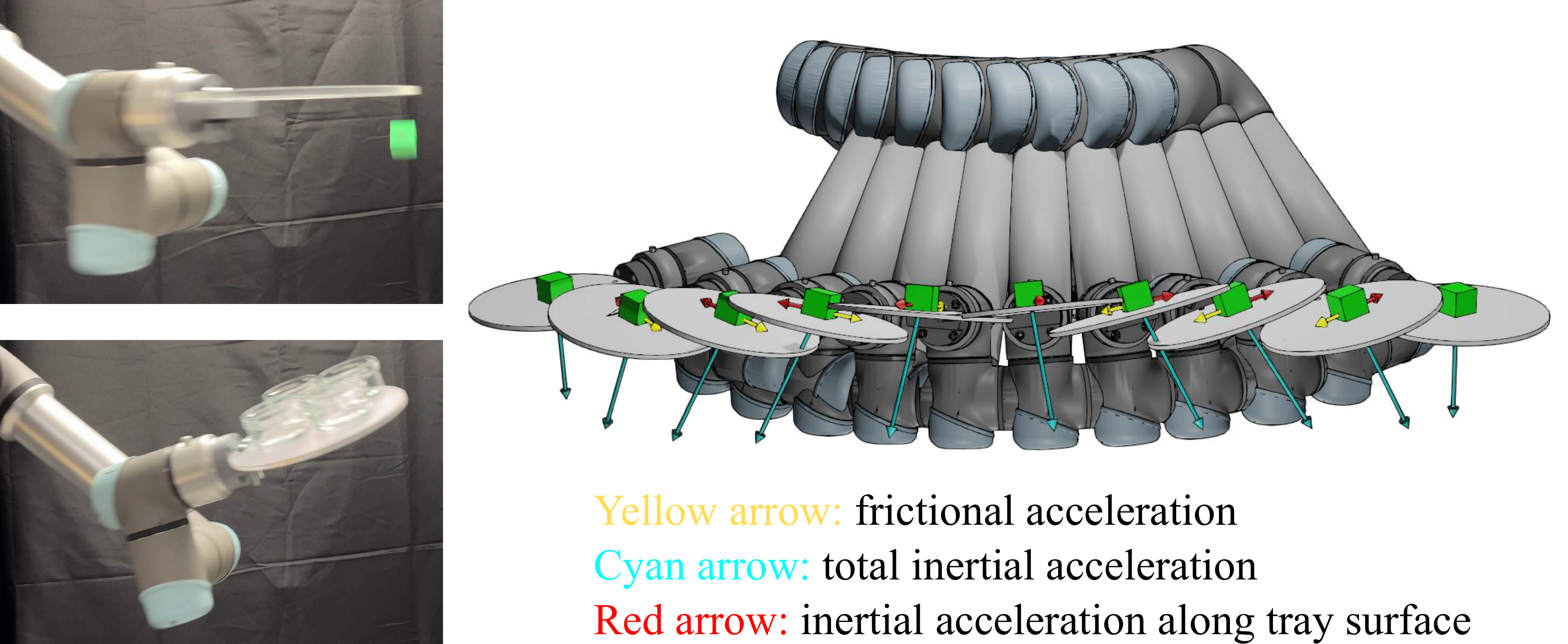}
    \caption{\textbf{Fast non-prehensile transport}. \textbf{Top left:} Time-optimized transport without constraints drops the object. \textbf{Bottom left:} With friction-based constraints, the motion planner initiates tray tilt to compensate for inertia effects, safely transporting multiple fragile objects. \textbf{Right:} We propose a learned dynamic constraint to balance the inertial and frictional forces during fast transport, reducing object sliding.}
    \label{fig:fig1}
    \vspace{-4mm    }
\end{figure}

Learning a dynamic friction model requires observing transported objects throughout high-speed motions. While camera-based setups are common in workcells, they require special hardware to have a high frame rate, avoid motion blur, and work around occlusions. While augmenting the tray with pressure-based tactile sensors would provide high-speed sensing that avoids occlusions, it comes at a significant additional expense and the potential for wear over time. Instead, we propose using acoustic sensing as it offers advantages including high sampling frequency, low latency, robustness to environmental noise, and ease of implementation, thus well-suited for detecting object sliding in non-prehensile transport. By attaching a contact microphone to the tray, we capture vibrations induced by object motion everywhere on the tray surface.

During high-speed motions constrained by the \emph{standard} Coulomb friction model, the attached contact microphone detects unexpected sliding motions correlated with the robot’s mechanical vibrations, which intensify with transport speed.
Leveraging the sensing data, we train a neural network to learn a friction coefficient that updates with varying tray velocity and acceleration, improving non-prehensile transport safety beyond limitations of methods based on the standard Coulomb friction model.

In experiments, we learn the dynamic friction model with data collected from a UR5e robot. We test constraints with the standard Coulomb friction model and the learned friction model on 8 objects with 12 different configurations. Results suggest that the learned friction model can reduce the object displacement by up to 86.0\,\% compared to the baseline.

The contributions of this paper are:

\begin{itemize}
    \item An acoustic sensing approach to learn effective friction between the tray and transported objects during high-speed non-prehensile transport.
    \item An integration of the learned friction model into a time-optimizing motion planner.
    \item Physical experiments on a UR5e robot suggesting the learned friction model lowers displacement over a standard Coulomb friction model during high-speed non-prehensile transport.
\end{itemize}

\section{Related Work}
\subsection{Non-Prehensile Transport}
The Waiter's Problem, proposed by Bernheisel and Lynch~\cite{bernheisel2004stable}, is the problem formulation most closely related to this work. It proposes transport of objects subject to unilateral constraints only and motion planning that takes into account inertial effects. Different approaches have been taken to secure objects on a moving tray~\cite{Ruggiero2018Survey}. Acharya et al.~\cite{acharya2020nonprehensile} focus on preventing object tipping and propose using s-curve motion profiles to control peak accelerations after analyzing the marginal stability of objects. Zhou et al.~\cite{Zhou2022DualArm} transports a tray with a dual-arm setup and employ MPC to compute grasping wrench, together with a force servoing controller to control contact forces between the tray and both end effectors. Selvaggio et al.~\cite{SelvaggioTCST2023} design an MPC framework to move the tray along a reference trajectory. Heins and Schoellig~\cite{Heins_2023} develop an MPC algorithm for a mobile manipulator to respond to dynamic obstacles while balancing the objects by optimizing trajectories online.

Unlike the previous works that primarily address the motion planning challenges of non-prehensile transport, our approach focuses on the performance of transport from the aspect of object displacement relative to the tray during transport. We push friction-based constraints to the real-world limits by addressing physical phenomena rather than relying solely on standard physics analysis of Coulomb friction. Instead of using minimum statically-feasible friction coefficients~\cite{Heins_2023} or a conservatively approximated friction cone space which may still lead to significant ($\geq$~1\,cm) object displacements during transport~\cite{SelvaggioTCST2023}, we measure the friction coefficients experimentally, use their exact values during motion planning for time optimization, and refine them with a learned model to boost performance. 

\subsection{Acoustic Sensing in Manipulation}
Acoustic sensing with contact microphones provides a low-cost solution for ensuring that signals are accurately aligned in time with little engineering effort, making data processing easier. In manipulation tasks, contact microphone data can facilitate perception and learning~\cite{gandhi2020swoosh}, \cite{pmlr-v229-thankaraj23a}, \cite{liu2024maniwav}. Clarke et al.~\cite{pmlr-v87-clarke18a} propose using vibrations from contact to measure flow and amount of granular materials during scooping and pouring using a signal processing method that converts audio into binned spectrograms. We adapt their approach to focus on detecting sudden increases in vibration intensity caused by object sliding, rather than training a model directly from the spectrograms' visual features.

\subsection{Motion Planning for Dynamic Manipulation}
Optimization-based motion planners generate smooth trajectories while enforcing constraints~\cite{ratliff2009chomp,kalakrishnan2011stomp,schulman2013finding}. In the context of dynamic manipulation, the kinematic and dynamic limitations of robots play an important role in the formulation of constraints to speed up motions~\cite{ichnowski2020gomp,ichnowski2020djgomp}. Additionally, the constraints should incorporate either analytical formulations of the problems~\cite{lynch1996dynamic,lynch1999dynamic,srinivasa2005using,ichnowski2022gompfit, pham2018fastsuction}, or learned models~\cite{zeng2020tossingbot,wang2020swingbot,zhang2021rotla, chi2022iterativeresidualpolicygoalconditioned, DBLP:conf/wafr/AvigalICG22}, to account for the dynamic properties during motion planning. Inspired by these prior works, we take an integrated approach for motion planning. We adapt the optimization-based motion planning framework from GOMP-FIT~\cite{ichnowski2022gompfit}, formulate a task-oriented constraint analytically, and embed a learned model in this constraint.

\section{Problem Statement} \label{problem_statement}
A robot with $n$ degrees of freedom has a tray end effector. Let $\mathrm{\bf{q}}\in \mathcal{C}$ be the angular configurations of all $n$ joints of the robot, where $\mathcal{C} \subset \mathbb{R}^n$ is the set of all valid configurations. Let $\mathrm{\bf{x}}_i = [\mathrm{\bf{q}}_i, \dot{\mathrm{\bf{q}}}_i, \ddot{\mathrm{\bf{q}}}_i] \in \mathcal{X}$ be the robot's dynamic state at any time step $i$. Let $\tau = \left(\mathrm{\bf{x}}_0, \mathrm{\bf{x}}_1, ... ,\mathrm{\bf{x}}_H\right)$ be the complete trajectory discretized to $H+1$ time steps. Let $f_k: \mathcal{C} \rightarrow SE(3)$ be the forward kinematics function that computes the pose of the tray.

Given the start and goal tray positions, $\mathrm{\bf{g}}_{\mathrm{start}},\mathrm{\bf{g}}_{\mathrm{goal}} \in SE(3)$, and the centroids of $m$ objects resting on the robot's tray, the objective is two-fold. First, identify task-oriented dynamic constraints $\mathcal{D}$ adaptable to real-world uncertainties and a function $f_c$ to map robot's state to the constrained parameters for each object to prevent sliding. Second, compute a time-optimized trajectory $\tau$ that satisfies the following conditions: $f_k(\mathrm{\bf{q}}_0)=\mathrm{\bf{g}}_{\mathrm{start}}$, $f_k(\mathrm{\bf{q}}_H)=\mathrm{\bf{g}}_{\mathrm{goal}}$, while ensuring that the robot kinematic and dynamic constraints, along with the identified constraints $\mathcal{D}$, are met at each time step.



For objects transported on a moving tray, we assume: 1) they have even contact surfaces, 2) they will not topple during transport, 3) all external forces are exerted at their centroids, and 4) they do not apply forces on each other. Additionally, we assume aerodynamic, adhesive, and electrostatic forces to be negligible during object transport.

\section{Preliminary}
We construct the motion planning framework for evaluating various friction models in fast non-prehensile transport. We formulate a baseline constraint with the standard Coulomb friction model and integrate it with the motion planner from GOMP-FIT~\cite{ichnowski2022gompfit} to solve the non-prehensile transport problem. This section provides background on GOMP-FIT, then discusses the integration of task-specific constraints.

\subsection{GOMP-FIT Background}
GOMP-FIT formulates motion planning for fast transport as an optimization problem and solves it with a sequential quadratic program (SQP) solver. GOMP-FIT discretizes the trajectory into a sequence of $H+1$ waypoints consisting of joint configurations and their derivatives, $\mathrm{\bf{x}}_i=\left[\mathrm{\bf{q}}_i, \dot{\mathrm{\bf{q}}}_i, \ddot{\mathrm{\bf{q}}}_i\right], i\in \left[0, \, H \right]$, each separated by a timestep $t_\mathrm{step}$. 

We employ the inner loop of GOMP-FIT that solves an SQP where the optimization is in the form:
\begin{align*}
    \arg\min_{\mathrm{\bf{x}}_{\left[0...H\right]}}\quad&\frac{1}{2}\mathrm{\bf{x}}^TP\mathrm{\bf{x}}+\mathrm{\bf{p}}^T\mathrm{\bf{x}} \\
    \text{s.t.} \quad & f_k(\mathrm{\bf{q}}_0)=\mathrm{\bf{g}}_{\mathrm{start}}, f_k(\mathrm{\bf{q}}_H)=\mathrm{\bf{g}}_{\mathrm{goal}} \\
    & \mathrm{\bf{q}}_i\in \mathcal{C} \quad &&\forall i \in \left[0, \, H\right]\\
    & \mathrm{\bf{q}}_i, \dot{\mathrm{\bf{q}}}_i, \ddot{\mathrm{\bf{q}}}_i, \dddot{\mathrm{\bf{q}}}_i \in \mathrm{joint\, limits} &&\forall i \in \left[0, \, H\right]\\
    & \mathrm{\bf{q}}_{i+1}=\mathrm{\bf{q}}_i+t_\mathrm{step}\dot{\mathrm{\bf{q}}}_i+\frac{1}{2}t^2\ddot{\mathrm{\bf{q}}}_i &&\forall i \in \left[0, \, H\right)\\
    & \dot{\mathrm{\bf{q}}}_{i+1}=\dot{\mathrm{\bf{q}}}_i+t_\mathrm{step}\ddot{\mathrm{\bf{q}}}_i &&\forall i \in \left[0, \, H\right)\\
    & \dddot{\mathrm{\bf{q}}}_{i+1} = \left(\ddot{\mathrm{\bf{q}}}_{i+1}-\ddot{\mathrm{\bf{q}}}_i \right)/t_\mathrm{step} &&\forall i \in \left[0, \, H\right)\\
    & f_c(\mathrm{\bf{q}}_i, \dot{\mathrm{\bf{q}}}_i, \ddot{\mathrm{\bf{q}}}_i) \in \mathcal{D} &&\forall i \in \left[0, \, H\right],
\end{align*}
where $P$ minimizes the sum-of-squared jerks, $\mathrm{\bf{p}}$ includes penalties for non-convex constraints, and $\mathcal{D}$ and $f_c$ together characterize the task-oriented constraints in dynamic manipulation. The SQP solver repeatedly linearizes the non-linear constraints to form a quadratic program (QP) and accepts the solutions that improve the trajectory. In the $(k+1)$ iterate, a non-linear constraint of the form $g(\mathrm{\bf{x}})\leq c$ is linearized around the current iterate $\mathrm{\bf{x}}^k$ through first-order approximation using its Jacobian: $J\mathrm{\bf{x}}^k+1 \leq J\mathrm{\bf{x}}^k-g(\mathrm{\bf{x}}^k)+c$.


We apply a time shrinking strategy similar to Curobo~\cite{curobo_report23}, keeping a fixed $H$ ($H{=}32$ in experiments) and shrinking $t_\mathrm{step}$ through binary search (starting from $t_\mathrm{step}=\text{0.25\,s}$ in experiments). The algorithm terminates when the SQP fails to find a solution or when the update of $t_\mathrm{step}$ between iterations is less than 50\,{\textmu}s. The planner returns the trajectory with the smallest $t_\mathrm{step}$ that allows the SQP to succeed as (locally) time-optimal. 


\subsection{Baseline Constraint with Standard Coulomb Friction Model} \label{constraint_static_mu}
The formulation of the baseline constraint is based on the friction and inertia at the centroid of each object. We derive the maximum static frictional force from the Coulomb Friction Model: $F_f \leq \mu_s F_n$, where $F_f$ and $F_n$ are the frictional force and normal force exerted by each contact surface on the other respectively, and $\mu_s$ is the static friction coefficient measured experimentally. We compute the acceleration due to gravitational, Euler, Coriolis, and centrifugal forces at each object's centroid, using the Recursive Newton Euler (RNE) method~\cite{luh1980line}.

When an object maintains complete contact with the tray, it remains stationary relative to the tray if the frictional force from the tray balances the inertial force tangential to the tray surface. Otherwise, it slides. We thus constrain trajectories to keep the inertial force along the tray surface below the maximum static frictional force. The constraint has the form:
\begin{equation}\label{ee_constraint_static_mu}
    \begin{split}
    &m\left\lVert \mu_s \left(\mathrm{\bf{a}}_i\cdot f_n(\mathrm{\bf{q}}_i)\right)\right\rVert \\ &- m\left\lVert\mathrm{\bf{a}}_{i}-\left(\mathrm{\bf{a}}_{i}\cdot f_n(\mathrm{\bf{q}}_i)\right)f_n(\mathrm{\bf{q}}_i)\right\rVert \geq 0, \quad \forall i \in \left[0, \,H\right],
    \end{split}
\end{equation}
where $m$ is the object mass, $\mathrm{\bf{a}}_{i}$ is the inertial acceleration at time step $i$, and $f_n(\mathrm{\bf{q}}_i)$ is the normal of the tray's contact surface at robot configuration $\mathrm{\bf{q}}_i$. 

\subsubsection*{One-Shot Measurement of Static Friction Coefficient}\label{measure_mu}
To derive the static friction coefficient, we assume only gravitational, normal, and frictional forces act on the object when the tray is stationary. An object begins to slide on the tray inclined at angle $\theta$ when the total force along the tray surface reaches equilibrium: $\mu_s (mg \cos{\theta}) - mg \sin{\theta} = 0 \Rightarrow \mu_s = \tan{\theta}$, where $m$ is the object mass and $g$ is gravity. Thus, we derive $\mu_s$ from the minimum $\theta$ that induces object sliding. To measure $\theta_\mathrm{min}$, we increment $\theta$ by 0.01 radians until the object starts to slide. In experiments, we measure $\mu_s$ for each material, and each measurement takes less than 1 minute.

\section{Method}\label{method}
Real-robot experiments with the baseline constraint show significant object displacement, indicating additional factors not characterized by the Coulomb friction model. Thus, we investigate the causes of object sliding during fast non-prehensile transport and propose a learning-based method to reduce object displacement.

\subsection{Motion Characterization with Acoustic Sensing}\label{motion_characterization}

We employ acoustic sensing with a passive piezoelectric microphone, known as a contact microphone, as it offers a direct and robust approach to detecting object displacement on a fast moving tray. When an object slides on the tray, it generates mechanical vibrations that propagate through the material. A contact microphone captures these vibrations from any point on the moving tray at a sampling frequency of 44.1\,kHz, creating a high-fidelity mapping between the physical event and the sensor output. 

\begin{figure}[t]
    \vspace{2mm}
    \centering
    \includegraphics[width=0.9\linewidth]{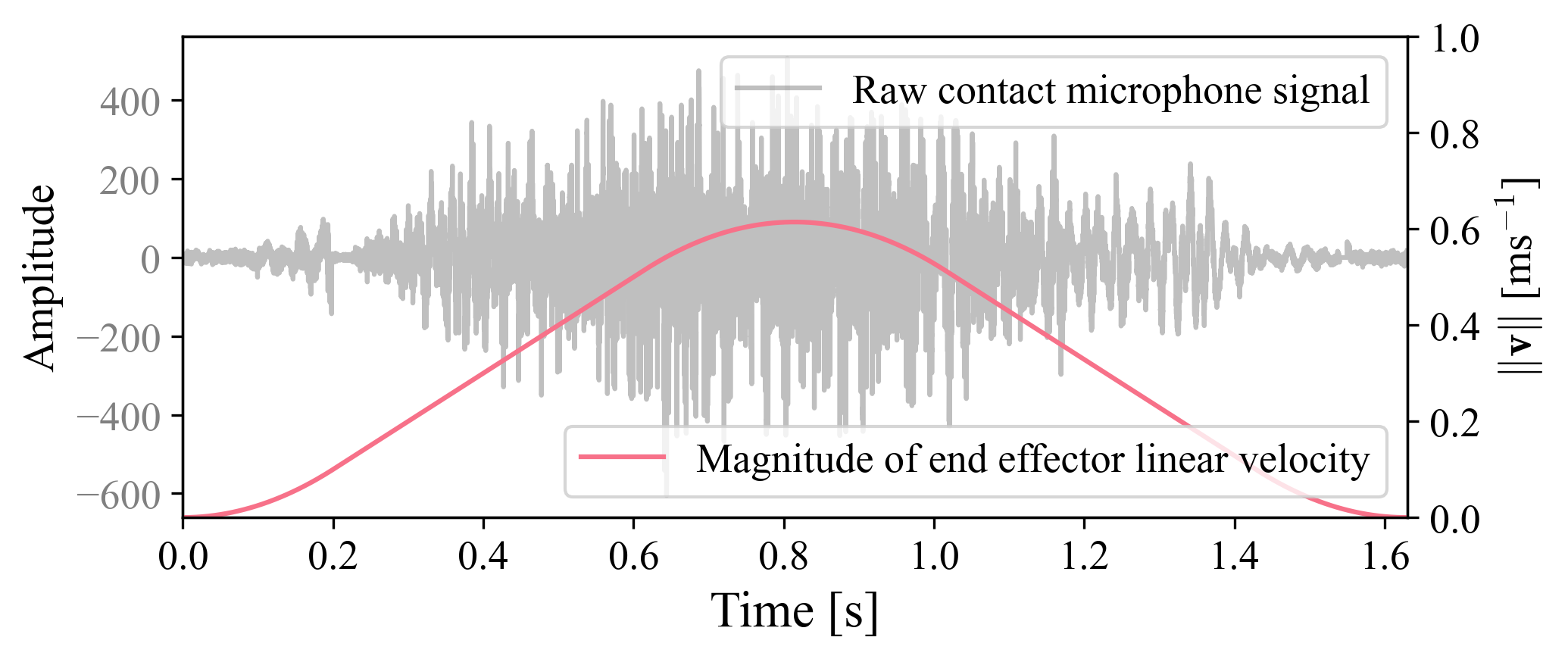}
    \caption{Raw contact-microphone signal collected when a UR5e robot's tray end effector accelerates at a maximum jerk of $\pm$4\,ms$^{-3}$, with no object on it. The overlay of acoustic signal amplitude and end-effector linear speed shows a positive correlation between the tray's vibration and speed.}
    \label{fig:vibration}
    \vspace{-4mm    }
\end{figure}

Experiments with a contact microphone attached to an empty tray show that tray vibrations intensify with the increase of the robot end-effector velocity (Fig.~\ref{fig:vibration}), even with low-jerk motions. These vibrations cause rapid fluctuations in the normal and tangential forces at the contact surfaces, resulting in uncertainty in the frictional force that holds objects in place. This suggests that factors contributing to object sliding that are not characterized by the standard Coulomb friction model become more significant as the tray speeds up. Thus, we propose learning a motion-aware friction model.

\subsection{Constraint with Learned Friction Model}
We use $\tilde{\mu}$, a dynamically conditioned friction coefficient, to take into account external forces not characterized by the static friction coefficient $\mu_s$, minimizing object sliding. The updated friction model has the form: $F_{t_{\mathrm{max}}} = \tilde{\mu} F_n$, where $F_n$ is the normal force and $F_{t_{\mathrm{max}}}$ is the maximum tangential force, both exerted on the object's contact surfaces by the tray. Here, $\tilde{\mu}$ changes with the robot's end-effector motion, but, at any moment in time, $\tilde{\mu}$ is constant anywhere on the tray. 

As tray vibration increases with velocity, and is highly correlated to sliding (Sec.~\ref{motion_characterization}), we train a dynamic parameter $\alpha$ for a specific robot to accommodate the vibrations inherent in its mechanical design such that $\tilde{\mu} = \alpha \mu_s$. The constraint from Eq.~\ref{ee_constraint_static_mu} now has the form:
\begin{equation} \label{constraint}
    \begin{split}
    &\left\lVert \alpha_i\mu_s \left(\mathrm{\bf{a}}_{i}\cdot f_n\left(\mathrm{\bf{q}}_i\right)\right)\right\rVert \\
    &- \left\lVert\mathrm{\bf{a}}_{i}-\left(\mathrm{\bf{a}}_{i}\cdot f_n(\mathrm{\bf{q}}_i)\right)f_n(\mathrm{\bf{q}}_i)\right\rVert \geq 0, \quad \forall i \in \left[0, \,H\right].
    \end{split}
\end{equation}

\subsection{Model Training with Acoustic Sensing} \label{mu_identification}

We apply a trial-and-error strategy during data collection to identify the threshold features when the object starts sliding. Similar to the GOMP-ST data-collection pipeline~\cite{DBLP:conf/wafr/AvigalICG22}, we use Ruckig~\cite{berscheid2021jerk} to plan a series of time-optimal straight-line horizontal motions for the tray between 2 fixed points, with varying acceleration and jerk limits. The robot keeps the tray level throughout the motion, with both linear velocity and acceleration aligned with the direction of movement. This alignment ensures consistent linear velocity and acceleration across the tray, eliminating the need to reposition the object after each motion. We then translate the motions into joint-space trajectories with an analytic inverse kinematics solver. We execute each trajectory 5 times without the object and 5 times with the object on the tray, recording contact microphone signals.

\begin{figure}[t]
    \vspace{2mm}
    \centering
    \includegraphics[width=0.9\linewidth]{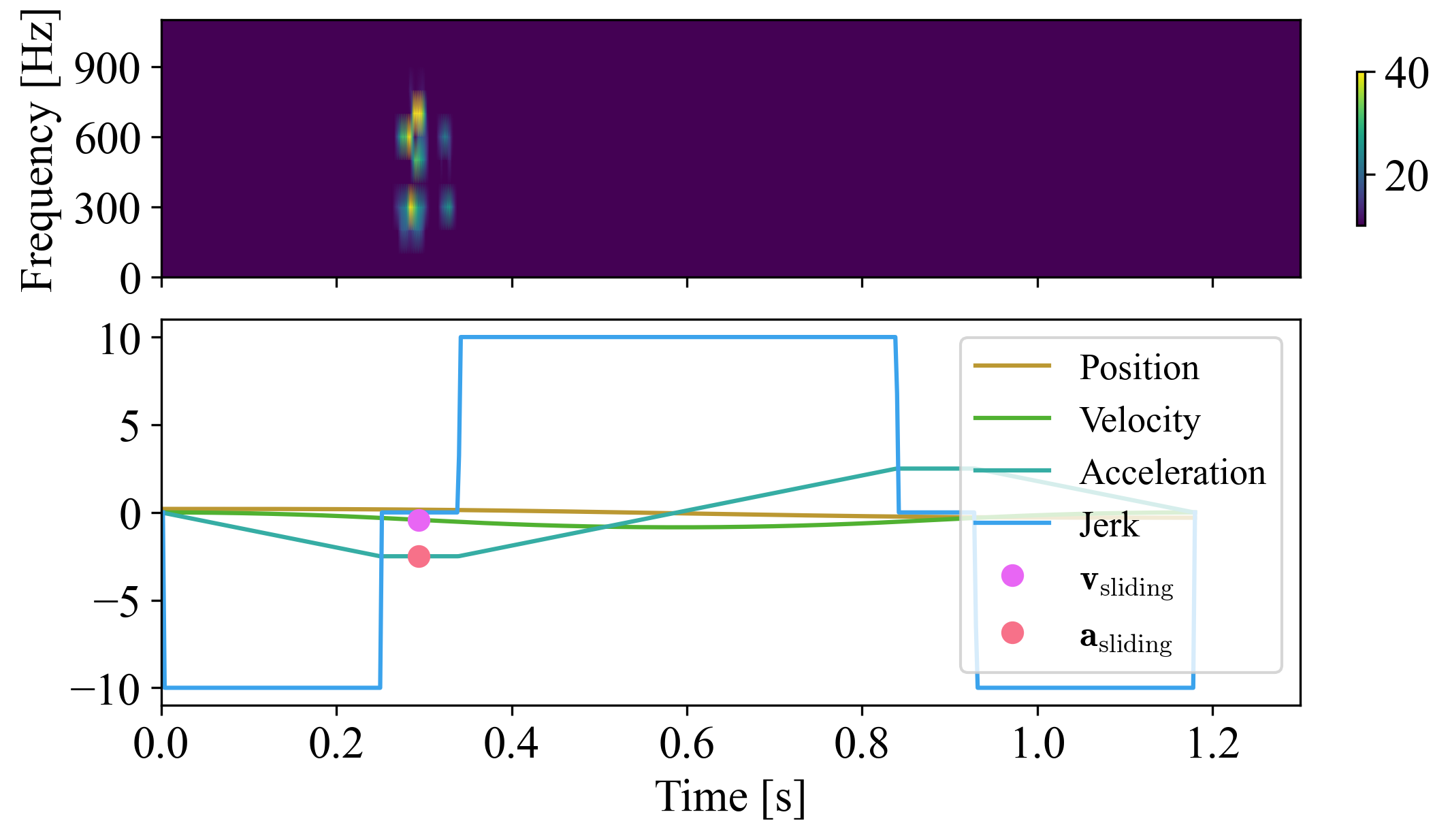}
    \caption{Synchronization of a fully preprocessed spectrogram with the planned trajectory in Cartesian space. We focus on the first occurrence of sliding, and thus mask out the part of spectrogram past the maximum velocity. The plot shows a trial with 
    a 0.5\,m straight-line horizontal end-effector trajectory with an acceleration limit of 2.5 ms$^{-2}$ and a jerk limit of 10 ms$^{-3}$. Here $t_\mathrm{sliding}=0.294$ s, $\lVert \mathrm{\bf{a}}_\mathrm{sliding} \rVert = 2.5$ ms$^{-2}$, $\lVert \mathrm{\bf{v}}_\mathrm{sliding} \rVert =0.423$ ms$^{-1}$.}
    \label{fig:feature_identification}
    \vspace{-4mm    }
\end{figure}

For data analysis, we use the Python audio package Librosa~\cite{mcfee2015librosa} and SciPy~\cite{2020SciPy-NMeth} to process raw signals and the Noisereduce Python package~\cite{tim_sainburg_2019_3243139} to extract signals representing the dynamic interactions between the object and the tray. In this process, we treat signals from trials with the object as the sound, and signals from trials without the object as the non-stationary noise~\cite{sainburg2020finding}. We then process the filtered signals into a binned spectrogram using the method from Clarke et al.~\cite{pmlr-v87-clarke18a}, with a time bin size matching the robot controller's time step and a frequency bin size of 100\,Hz. We identify the time $t_\mathrm{sliding}$ when the signal first shows a significantly greater magnitude and correlate it with the velocity $\mathrm{\bf{v}}_\mathrm{sliding}$ and acceleration $\mathrm{\bf{a}}_\mathrm{sliding}$ in the planned Cartesian-space trajectory at that specific time step (Fig.~\ref{fig:feature_identification}). These pairs of  $\mathrm{\bf{v}}_\mathrm{sliding}$ and $\mathrm{\bf{a}}_\mathrm{sliding}$ are identified as threshold features of object sliding. 

After data analysis, we further process $\mathrm{\bf{a}}_\mathrm{sliding}$ into $\alpha_\mathrm{training}$ with the assumption that horizontal inertial force and gravitational force are the only forces exerted on the object during the straight-line horizontal motions:
\begin{equation*}
    m\left\lVert \mathrm{\bf{a}}_\mathrm{sliding} \right\lVert = m\alpha_\mathrm{training}\mu_s\left\lVert \mathrm{\bf{g}} \right\lVert \;\Rightarrow\; \alpha_\mathrm{training} = \frac{\left\lVert \mathrm{\bf{a}}_\mathrm{sliding} \right\lVert}{\mu_s\left\lVert \mathrm{\bf{g}} \right\lVert},
\end{equation*}
%
%
where $\mathrm{\bf{g}}$ is the gravity, $m$ is the object mass, and $\mu_s$ is the measured static friction coefficient. We then train a function $\alpha = g(\lVert \mathrm{\bf{v}} \rVert)$ that computes the motion-aware friction coefficient, where $\mathrm{\bf{v}}$ is the tray's linear velocity. Therefore, we formulate the complete constraint from Eq.~\ref{constraint} as:
\begin{align*}
    & \left\lVert \mu_s g(f_{\lVert v \rVert}(\mathrm{\bf{q}}_i, \dot{\mathrm{\bf{q}}}_i, \ddot{\mathrm{\bf{q}}}_i))\left[f_a(\mathrm{\bf{q}}_i, \dot{\mathrm{\bf{q}}}_i, \ddot{\mathrm{\bf{q}}}_i)\cdot f_n(\mathrm{\bf{q}}_i)\right] \right\rVert \\
    & - \left\lVert f_a(\mathrm{\bf{q}}_i, \dot{\mathrm{\bf{q}}}_i, \ddot{\mathrm{\bf{q}}}_i) - \left[f_a(\mathrm{\bf{q}}_i, \dot{\mathrm{\bf{q}}}_i, \ddot{\mathrm{\bf{q}}}_i) \cdot f_n(\mathrm{\bf{q}}_i)\right]f_n(\mathrm{\bf{q}}_i) \right\rVert \geq 0,  \\
    &\forall i \in [0, H],
\end{align*}
where $\mu_s$ is the measured static friction coefficient from Sec.~\ref{measure_mu}, $g: \mathbb{R} \rightarrow \mathbb{R}$ is the learned function that computes the dynamic constraint parameter $\alpha$ from the magnitude of linear velocity, $f_{\lVert v \rVert}: \mathcal{X} \rightarrow \mathbb{R}$ computes the magnitude of linear velocity, and $f_a: \mathcal{X} \rightarrow \mathbb{R}^{3}$ computes the linear inertial acceleration with RNE method~\cite{luh1980line}. $f_n: \mathbb{R}^n \rightarrow \mathbb{R}^3$ computes the normal vector of the tray attached to a robot arm with $n$ degrees of freedom, which is the same anywhere on the tray. In implementation, we vectorize the constraint to compute for multiple transported objects in parallel.


\section{Experiments}
We conduct real-world experiments on a UR5e robotic manipulator with a custom tray as its end effector. We measure $\mu_s$ for each tested material, collect data and train dynamic friction models using 2 different objects, and then generate trajectories to transport objects.


\begin{figure}[t]
    \vspace{2mm}
    \centering
    \includegraphics[width=0.8\linewidth]{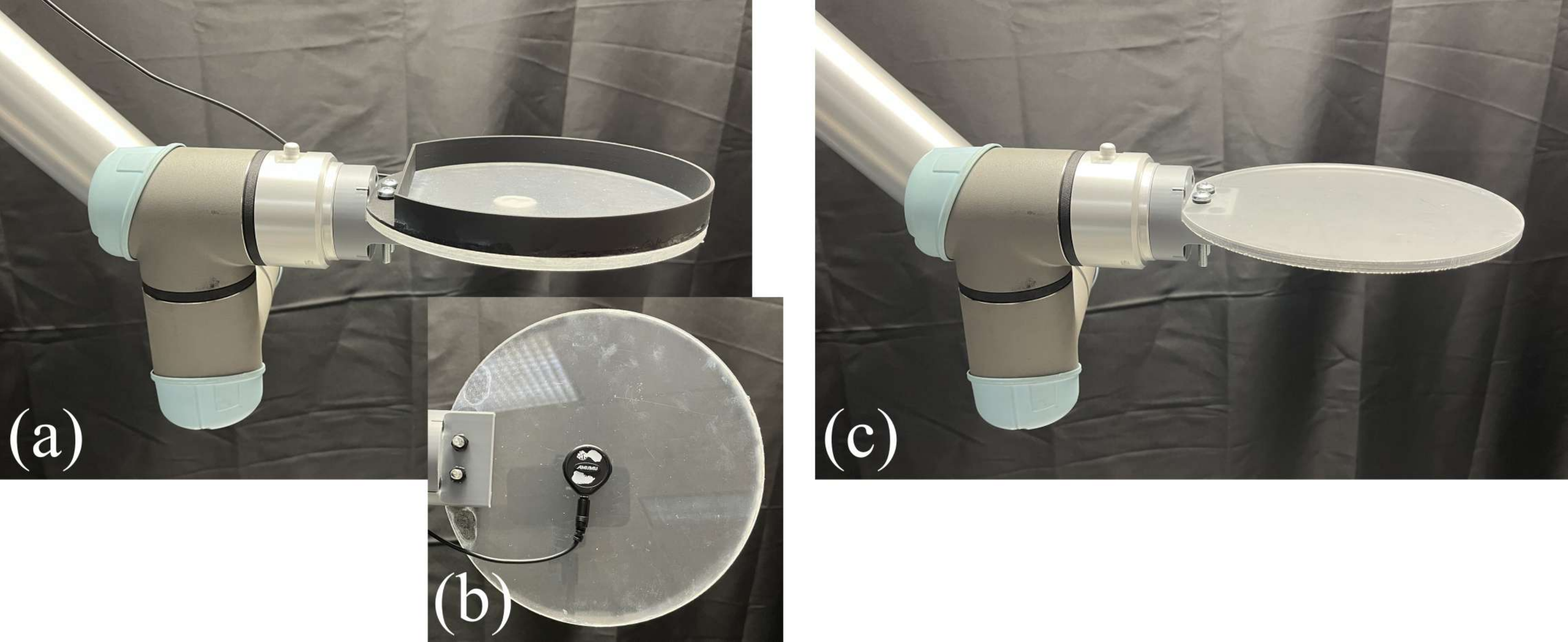}
    \caption{(a) Data collection setup. (b) Contact microphone attached under the tray. (c) Model evaluation setup. We leave the protective film on the tray to prevent the electrostatic friction caused by acrylic. }
    \label{fig:plate}
    \vspace{-4mm    }
\end{figure}

\subsection{Model Training}

During data collection, we use a round acrylic sheet with a 3D-printed PLA boundary as the tray (Fig.~\ref{fig:plate}a). This setup allows for automatic data collection without manual resets as the boundary prevents objects from falling. We attach a AMUMU SBT-10 Trans-HD Transducer at the center of the tray's bottom surface (Fig.~\ref{fig:plate}b) and use Lavalier USB UL20 to amplify the signals and convert them to USB signals.

\begin{figure}[t]
    \vspace{1mm}
    \centering
    \includegraphics[width=0.9\linewidth]{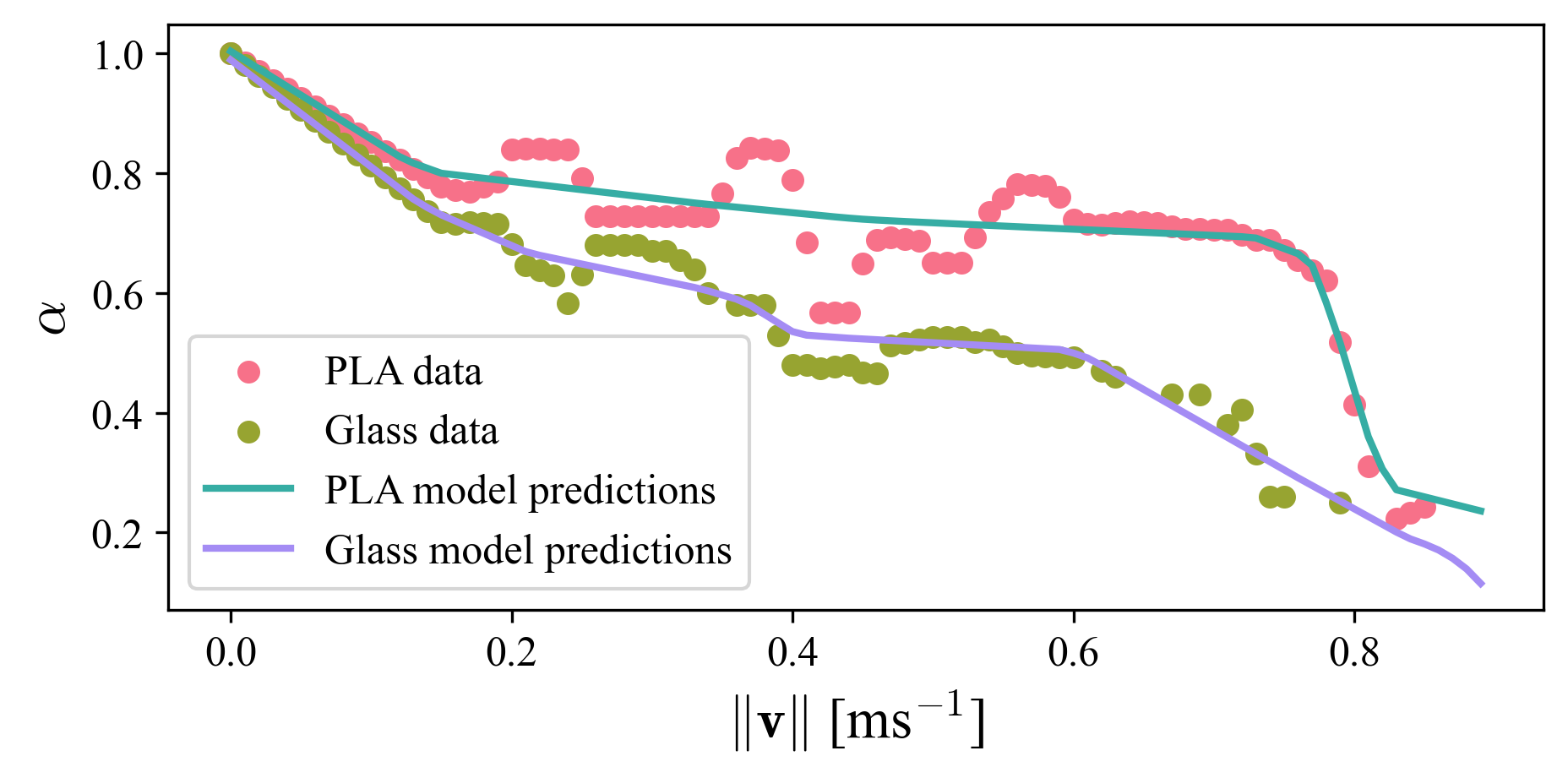}
    \caption{Visualization of the models trained for dynamic constraint parameter $\alpha$.}
    \label{fig:model_viz}
    \vspace{-4mm}
\end{figure}

We conduct data collection with a 3D-printed PLA cube (Fig.~\ref{fig:objects}a) and a small glass container (Fig.~\ref{fig:objects}c). We generate straight-line horizontal trajectories between the start position $\left(0.20, 0.60, 0.20\right)$ and the end position $\left(-0.30, 0.60, 0.20\right)$ in meters relative to the robot base frame, using Ruckig with acceleration limits in [1.0, 2.5]\,ms$^{-2}$ and jerk limits in [1.0, 10.0]\,ms$^{-3}$. We identify 85 unique pairs of $\lVert\mathrm{\bf{v}}_\mathrm{sliding}\rVert$ and $\lVert\mathrm{\bf{a}}_\mathrm{sliding}\rVert$ from the PLA cube and 70 unique pairs from the glass container. We augment the data by assuming $\lVert\mathrm{\bf{a}}_\mathrm{sliding}\lVert = \mu_s\lVert\mathrm{\bf{g}}\rVert$ when $\lVert\mathrm{\bf{v}}_\mathrm{sliding}\rVert = 0$ and perform linear interpolation to the minimum identified $\lVert\mathrm{\bf{v}}_\mathrm{sliding}\rVert$. We then train an MLP model for $\alpha$ using the data from each object, consisting of 3 fully connected layers with LeakyReLU activations and dropouts for regularization. Fig.~\ref{fig:model_viz} shows the models that reduce $\alpha$ as velocity increases to compensate for the impact of forces under stronger vibration, with the glass function being more conservative. We test each model separately to evaluate the performance.


\subsection{Model Evaluation}


\begin{figure}[t]
    \centering
    \includegraphics[width=0.8\linewidth]{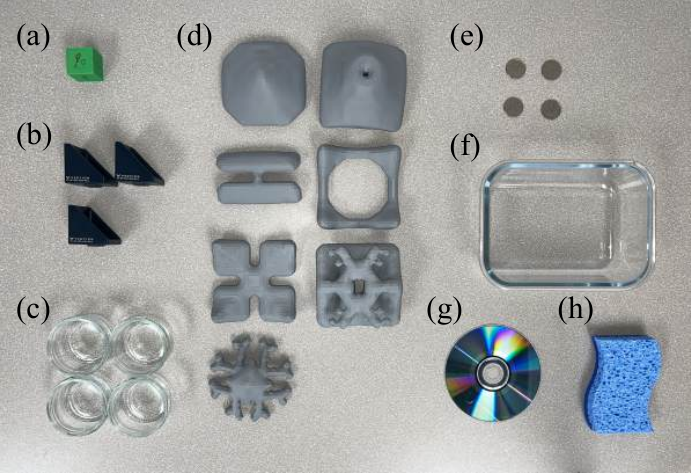}
    \caption{\textbf{Objects transported in experiments.} (a) A 3D printed PLA cube: light objects with a smooth contact surface. (b) Aluminum gussets: heavy objects with a smooth contact surface. (c) Small glass containers: fragile objects. (d) EGAD! objects printed with PLA: large, irregularly-shaped objects. (e) US quarter coins: small, thin objects. (f) A large glass container: large, heavy, fragile objects. (g) A CD-R disc: large, thin objects. (h) A sponge: lightweight, deformable, and porous objects with an uneven contact surface.}
    \label{fig:objects}
    \vspace{-4mm    }
\end{figure}

\begin{table*}[t]
\vspace{2mm}
\centering
\caption{\textbf{Mean object displacement [mm] after transport.}} 
\label{tab:displacement}
\scriptsize
\begin{tabular}{@{}l@{\quad}c@{\quad}c@{\quad}c@{\quad}c@{\quad}c@{\quad}c@{\quad}c@{\quad}c@{\quad}c@{\quad}c@{\quad}c@{\quad}c@{}}
\toprule
& \makecell{Cube \\ $\times 1$} & \makecell{Gusset \\ $\times 1$} & \makecell{Glass small \\ $\times 1$} & \makecell{EGAD!$^\mathrm{(a)}$\\ $\times 1$} & \makecell{Coin \\ $\times 1$} & \makecell{Glass big \\ $\times 1$} & \makecell{Gusset \\ $\times 3$} & \makecell{Coin \\ $\times 4$} & \makecell{Glass small \\ $\times 4$} & \makecell{Disc \\ $\times 1$} & \makecell{Sponge$^\mathrm{(a)}$\\ hard side} & \makecell{Sponge$^\mathrm{(a)}$\\ soft side}\\
\midrule
Material & PLA & Aluminum & Glass & PLA & Cupronickel & Glass & Aluminum & Cupronickel & Glass & Polycarbonate & Polymer & Foam \\
$\mu_s$ & 0.21 & 0.18 & 0.17 & 0.21 & 0.21 & 0.17 & 0.18 & 0.21 & 0.17 & 0.30 & 0.21  & 0.25 \\ \midrule

Coulomb & 18.3 & 15.9 & 15.3 & 18.9 & 14.7 & 8.2 & 16.7 & 18.3 & 13.2 & 15.1 & 21.2 & 13.6 \\
Learned-PLA & 8.1 & 6.1 & 2.8 & 9.8 & 7.6 & 2.6 & 9.0 & 12.4 & 3.1 & 5.3 & 13.7 & 9.9 \\
Learned-Glass & 2.7 & 1.5 & 1.4 & 3.1 & 1.9 & 1.7 & 4.6 & 1.8 & 1.5 & 1.4 & 2.9 & 1.7 \\
AS-Coulomb & 3.9 & 2.7 & 2.1 & 4.4 & 3.2 & 2.3 & 4.6 & 3.6 & 2.3 & 3.8 & 5.8 & 1.1 \\
AS-No Constraints & Fail & Fail & Fail & Fail & 35.2 & -$^{*}$ & Fail & 26.5 & Fail & 21.2 & 38.5 & 15.6 \\

\bottomrule

\end{tabular} \\
\scriptsize Assumption violations: 
$\mathrm{(a)}$uneven contact surface \qquad
* Not tested for safety reasons. 
\end{table*}
\begin{table*}[t]
\centering
\caption{\textbf{Time [s] to transport objects.}}
\label{tab:time}
\scriptsize
\begin{tabular}{@{}l@{\qquad}c@{\quad}c@{\quad}c@{\quad}c@{\quad}c@{\quad}c@{\quad}c@{\quad}c@{\quad}c@{\quad}c@{\quad}c@{\quad}c@{}}
\toprule
& \makecell{Cube \\ $\times 1$} & \makecell{Gusset \\ $\times 1$} & \makecell{Glass small \\ $\times 1$} & \makecell{EGAD! \\ $\times 1$} & \makecell{Coin \\ $\times 1$} & \makecell{Glass big \\ $\times 1$} & \makecell{Gusset \\ $\times 3$} & \makecell{Coin \\ $\times 4$} & \makecell{Glass small \\ $\times 4$} & \makecell{Disc \\ $\times 1$} & \makecell{Sponge \\ hard side} & \makecell{Sponge \\ soft side}\\
\midrule


Coulomb & 1.124 & 1.160 & 1.174 & 1.124 & 1.124 & 1.174 & 1.200 & 1.176 & 1.226 & 1.058 & 1.124 & 1.086 \\
Learned-PLA &  1.212 & 1.272 & 1.382 & 1.212 & 1.212 & 1.382 & 1.282 & 1.316 & 1.406 & 1.396 & 1.212 & 1.188\\
Learned-Glass, AS & 1.406 & 1.500 & 1.516 & 1.406 & 1.406 & 1.516 & 1.526 & 1.500 & 1.525 & 1.372 & 1.406 & 1.376 \\

\bottomrule

\end{tabular} \\

\end{table*}

We use an acrylic sheet without boundary (Fig.~\ref{fig:plate}c) during constraint evaluation experiments, so the objects will fall if the constraint fails to keep them in place. We mount a top-down camera above the tray's target position to capture objects' final positions after each motion. We attach a sticker with 5$\times$5\,mm$^2$ grid patterns at the edge of the tray's top surface as a reference for real-word distance when computing object displacements using a computer vision algorithm. We select 8 objects (Fig.~\ref{fig:objects}) and conduct 12 sets of experiments by varying object configurations (single vs. multiple) and contact surfaces (hard vs. soft for Fig.~\ref{fig:plate}h). Particularly, we choose objects with maximum grasp difficulty from the evaluation set of EGAD!\cite{morrison2020egadevolvedgraspinganalysis} (Fig.~\ref{fig:objects}d) to demonstrate the tray's capability of handling irregular-shaped objects. And we 3D print them with PLA to test whether complex geometries affect the constraint performance when comparing with the PLA cube (Fig.~\ref{fig:objects}a). 

We first measure $\mu_s$ for each material and the positions of objects' centroids when placed on the tray (either a single object at the center or multiple objects evenly around the center), then input these values into the optimizer to generate trajectories for each object. The optimizer sets the start and end position at (0.75, 0.30, 0.20) and (0.15, 0.80, 0.20) respectively, in meters relative to the robot base frame, and generates arced trajectories to move the tray. In contrast to the short, straight-line trajectories used during data collection, these longer arced trajectories test whether the constraint properly accounts for centrifugal and Coriolis forces. They also enable the robot to achieve higher acceleration within the same jerk limits, allowing us to evaluate the performance during faster transport. Additionally, maintaining the same start and end height lets us observe whether constraints with the learned friciton model allow the motion planner to optimize speed by tilting the tray~\cite{bernheisel2004stable}.

For each object, we generate 1 time-optimized trajectory without constraints and 3 distinct constrained time-optimized trajectories: 1 with the standard Coulomb friction model (Coulomb), 2 with the learned friction model, trained with the PLA data (Learned-PLA) and the glass data (Learned-Glass) respectively. As the glass model is more conservative and results in a slower motion that takes $t_\mathrm{glass}$ seconds, we also generate 2 other trajectories for ablation study to verify the model is not just learning to go slower: 1 without constraints (AS-No Constraints) and 1 with the standard Coulomb friction constraint (AS-Coulomb), for both we terminate time optimization at $t_\mathrm{glass}$ seconds.

Experiments where objects fall off the tray are failure cases. On success, we measure object displacements relative to the tray in millimeters, by analyzing the top-down photos of the objects' initial and final positions with Segment Anything (SAM)~\cite{kirillov2023segany}. In experiments involving multiple objects on the tray, the displacement of each trial is defined as the average displacement of all centroids in that trial. For the 7 EGAD! objects, we transport each of them once and compute the average displacement across the 7 trials. We transport each remaining object 5 times and take the average displacement from these 5 trials.

\subsection{Results} \label{results}
Table~\ref{tab:displacement} shows the average displacements in millimeters. Table~\ref{tab:time} shows the duration of each trajectory in seconds. 

All time-optimized trajectories without constraints result in failures, so we do not record them in the tables. Objects remain on the tray during transport with constraint using the standard Coulomb friction model, suggesting that the baseline constraint reflects some, but not all of the real-world effects. All objects transported with constraints using the learned friction model show improved performance, with object displacement reduced by an average of 53.9\,\% and 86.0\,\% with the PLA model and the more conservative glass model respectively. In ablation study AS-No Constraints, 6 out of the 11 tested objects slide off the tray, while the remaining exhibit significant displacement. Except for the 2 outliers, AS-Coulomb trajectories perform worse in all cases, with displacements up to $2.7\times$ those from the learned constraint using the glass function. Among the 2 outliers, the gussets exhibit the same displacements using both trajectories, and the sponge is considered an out-of-distribution object due to its uneven contact surface: the hard side contacts only in the center, while the soft side touches at the edges. The ablation study indicates that the learned friction model captures phenomena beyond those accounted for by the standard Coulomb friction model, rather than scaling the measured friction coefficient by a constant $\alpha <1$ to reduce transport speed.

For the comparison of speed, the learned friction model increases the transport time by an average of 12.6\,\% and 27.0\,\% using the PLA function and the glass function respectively. This suggests an opportunity for practitioners to make a trade-off between speed and displacement.

\section{Conclusion and Future Work}
We propose learning a dynamic friction model with acoustic sensing to account for factors not characterized by the standard Coulomb friction model, and integrating it into a time-optimizing motion planner. We test on a UR5e robot with a tray as the end effector. Experimental results suggest that the learned friction model reduces object displacements after transport up to 86.0\,\% compared to constraining motions with the standard Coulomb friction model, making the proposed method practical in many real-world automation applications.


We consider 2 future directions for this work. First, the current tray design with a contact microphone is less effective at capturing vibrations from lightweight objects like a sponge. We will optimize the tray design and explore alternative piezoelectric sensors to improve sensing sensitivity. Second, we aim to extend friction modeling with acoustic sensing to more complex tasks, such as dexterous manipulation using higher-DOF end effectors.

\bibliographystyle{IEEEtran}
\bibliography{refs}

\begin{thebibliography}{10}
\providecommand{\url}[1]{#1}
\csname url@samestyle\endcsname
\providecommand{\newblock}{\relax}
\providecommand{\bibinfo}[2]{#2}
\providecommand{\BIBentrySTDinterwordspacing}{\spaceskip=0pt\relax}
\providecommand{\BIBentryALTinterwordstretchfactor}{4}
\providecommand{\BIBentryALTinterwordspacing}{\spaceskip=\fontdimen2\font plus
\BIBentryALTinterwordstretchfactor\fontdimen3\font minus \fontdimen4\font\relax}
\providecommand{\BIBforeignlanguage}[2]{{%
\expandafter\ifx\csname l@#1\endcsname\relax
\typeout{** WARNING: IEEEtran.bst: No hyphenation pattern has been}%
\typeout{** loaded for the language `#1'. Using the pattern for}%
\typeout{** the default language instead.}%
\else
\language=\csname l@#1\endcsname
\fi
#2}}
\providecommand{\BIBdecl}{\relax}
\BIBdecl

\bibitem{mahler2016dex}
J.~Mahler, F.~T. Pokorny, B.~Hou, M.~Roderick, M.~Laskey, M.~Aubry, K.~Kohlhoff, .~T. Kr{\"o}ger, J.~Kuffner, and K.~Goldberg, ``{Dex-Net} 1.0: A cloud-based network of {3D} objects for robust grasp planning using a multi-armed bandit model with correlated rewards.''\hskip 1em plus 0.5em minus 0.4em\relax IEEE, 2016, pp. 1957--1964.

\bibitem{wang2019adversarial}
D.~Wang, D.~Tseng, P.~Li, Y.~Jiang, M.~Guo, M.~Danielczuk, J.~Mahler, J.~Ichnowski, and K.~Goldberg, ``Adversarial grasp objects,'' in \emph{2019 IEEE 15th International Conference on Automation Science and Engineering (CASE)}, 2019, pp. 241--248.

\bibitem{morrison2020egadevolvedgraspinganalysis}
\BIBentryALTinterwordspacing
D.~Morrison, P.~Corke, and J.~Leitner, ``Egad! an evolved grasping analysis dataset for diversity and reproducibility in robotic manipulation,'' 2020. [Online]. Available: \url{https://arxiv.org/abs/2003.01314}
\BIBentrySTDinterwordspacing

\bibitem{mahler2018dex}
J.~Mahler, M.~Matl, X.~Liu, A.~Li, D.~Gealy, and K.~Goldberg, ``{Dex-Net} 3.0: Computing robust vacuum suction grasp targets in point clouds using a new analytic model and deep learning.''\hskip 1em plus 0.5em minus 0.4em\relax IEEE, 2018, pp. 1--8.

\bibitem{DBLP:conf/wafr/AvigalICG22}
\BIBentryALTinterwordspacing
Y.~Avigal, J.~Ichnowski, M.~Y. Cao, and K.~Goldberg, ``{GOMP-ST:} grasp optimized motion planning for suction transport,'' in \emph{Algorithmic Foundations of Robotics {XV} - Proceedings of the Fifteenth Workshop on the Algorithmic Foundations of Robotics, {WAFR} 2022, College Park, MD, USA, 22-24 June, 2022}, ser. Springer Proceedings in Advanced Robotics, S.~M. LaValle, J.~M. O'Kane, M.~W. Otte, D.~Sadigh, and P.~Tokekar, Eds., vol.~25.\hskip 1em plus 0.5em minus 0.4em\relax Springer, 2022, pp. 488--505. [Online]. Available: \url{https://doi.org/10.1007/978-3-031-21090-7\_29}
\BIBentrySTDinterwordspacing

\bibitem{agboh2022multiobjectgraspingplane}
\BIBentryALTinterwordspacing
W.~C. Agboh, J.~Ichnowski, K.~Goldberg, and M.~R. Dogar, ``Multi-object grasping in the plane,'' 2022. [Online]. Available: \url{https://arxiv.org/abs/2206.00229}
\BIBentrySTDinterwordspacing

\bibitem{agboh2023learningefficientlyplanrobust}
\BIBentryALTinterwordspacing
W.~C. Agboh, S.~Sharma, K.~Srinivas, M.~Parulekar, G.~Datta, T.~Qiu, J.~Ichnowski, E.~Solowjow, M.~Dogar, and K.~Goldberg, ``Learning to efficiently plan robust frictional multi-object grasps,'' 2023. [Online]. Available: \url{https://arxiv.org/abs/2210.07420}
\BIBentrySTDinterwordspacing

\bibitem{SelvaggioTCST2023}
M.~Selvaggio, A.~Garg, F.~Ruggiero, G.~Oriolo, and B.~Siciliano, ``Non-prehensile object transportation via model predictive non-sliding manipulation control,'' \emph{IEEE Transactions on Control Systems Technology}, pp. 1--14, 2023.

\bibitem{Heins_2023}
\BIBentryALTinterwordspacing
A.~Heins and A.~P. Schoellig, ``Keep it upright: Model predictive control for nonprehensile object transportation with obstacle avoidance on a mobile manipulator,'' \emph{IEEE Robotics and Automation Letters}, vol.~8, no.~12, p. 7986–7993, Dec. 2023. [Online]. Available: \url{http://dx.doi.org/10.1109/LRA.2023.3324520}
\BIBentrySTDinterwordspacing

\bibitem{bernheisel2004stable}
J.~D. Bernheisel and K.~M. Lynch, ``Stable transport of assemblies: Pushing stacked parts,'' \emph{IEEE Transactions on Automation science and Engineering}, vol.~1, no.~2, pp. 163--168, 2004.

\bibitem{Ruggiero2018Survey}
F.~Ruggiero, V.~Lippiello, and B.~Siciliano, ``Nonprehensile dynamic manipulation: A survey,'' \emph{IEEE Robotics and Automation Letters}, vol.~3, no.~3, pp. 1711--1718, 2018.

\bibitem{acharya2020nonprehensile}
P.~Acharya, K.-D. Nguyen, H.~M. La, D.~Liu, and I.-M. Chen, ``Nonprehensile manipulation: a trajectory-planning perspective,'' \emph{IEEE/ASME Transactions on Mechatronics}, vol.~26, no.~1, pp. 527--538, 2020.

\bibitem{Zhou2022DualArm}
C.~Zhou, M.~Lei, L.~Zhao, Z.~Wang, and Y.~Zheng, ``Topp-mpc-based dual-arm dynamic collaborative manipulation for multi-object nonprehensile transportation,'' in \emph{2022 International Conference on Robotics and Automation (ICRA)}, 2022, pp. 999--1005.

\bibitem{gandhi2020swoosh}
D.~Gandhi, A.~Gupta, and L.~Pinto, ``Swoosh! rattle! thump! -- actions that sound,'' 2020.

\bibitem{pmlr-v229-thankaraj23a}
\BIBentryALTinterwordspacing
A.~Thankaraj and L.~Pinto, ``That sounds right: Auditory self-supervision for dynamic robot manipulation,'' in \emph{Proceedings of The 7th Conference on Robot Learning}, ser. Proceedings of Machine Learning Research, J.~Tan, M.~Toussaint, and K.~Darvish, Eds., vol. 229.\hskip 1em plus 0.5em minus 0.4em\relax PMLR, 06--09 Nov 2023, pp. 1036--1049. [Online]. Available: \url{https://proceedings.mlr.press/v229/thankaraj23a.html}
\BIBentrySTDinterwordspacing

\bibitem{liu2024maniwav}
Z.~Liu, C.~Chi, E.~Cousineau, N.~Kuppuswamy, B.~Burchfiel, and S.~Song, ``Maniwav: Learning robot manipulation from in-the-wild audio-visual data,'' \emph{arXiv preprint arXiv:2406.19464}, 2024.

\bibitem{pmlr-v87-clarke18a}
\BIBentryALTinterwordspacing
S.~Clarke, T.~Rhodes, C.~G. Atkeson, and O.~Kroemer, ``Learning audio feedback for estimating amount and flow of granular material,'' in \emph{Proceedings of The 2nd Conference on Robot Learning}, ser. Proceedings of Machine Learning Research, A.~Billard, A.~Dragan, J.~Peters, and J.~Morimoto, Eds., vol.~87.\hskip 1em plus 0.5em minus 0.4em\relax PMLR, 29--31 Oct 2018, pp. 529--550. [Online]. Available: \url{https://proceedings.mlr.press/v87/clarke18a.html}
\BIBentrySTDinterwordspacing

\bibitem{ratliff2009chomp}
N.~Ratliff, M.~Zucker, J.~A. Bagnell, and S.~Srinivasa, ``{CHOMP}: Gradient optimization techniques for efficient motion planning,'' in \emph{2009 IEEE International Conference on Robotics and Automation}.\hskip 1em plus 0.5em minus 0.4em\relax IEEE, 2009, pp. 489--494.

\bibitem{kalakrishnan2011stomp}
M.~Kalakrishnan, S.~Chitta, E.~Theodorou, P.~Pastor, and S.~Schaal, ``{STOMP}: Stochastic trajectory optimization for motion planning,'' in \emph{2011 IEEE international conference on robotics and automation}.\hskip 1em plus 0.5em minus 0.4em\relax IEEE, 2011, pp. 4569--4574.

\bibitem{schulman2013finding}
J.~Schulman, J.~Ho, A.~X. Lee, I.~Awwal, H.~Bradlow, and P.~Abbeel, ``Finding locally optimal, collision-free trajectories with sequential convex optimization.'' in \emph{Robotics: Science and Systems}, 2013, pp. 1--10.

\bibitem{ichnowski2020gomp}
J.~Ichnowski, M.~Danielczuk, J.~Xu, V.~Satish, and K.~Goldberg, ``{GOMP}: Grasp-optimized motion planning for bin picking,'' in \emph{2020 International Conference on Robotics and Automation (ICRA)}.\hskip 1em plus 0.5em minus 0.4em\relax IEEE, May 2020.

\bibitem{ichnowski2020djgomp}
J.~Ichnowski, Y.~Avigal, V.~Satish, and K.~Goldberg, ``Deep learning can accelerate grasp-optimized motion planning,'' \emph{Science Robotics}, vol.~5, no.~48, 2020.

\bibitem{lynch1996dynamic}
K.~M. Lynch and M.~T. Mason, ``Dynamic underactuated nonprehensile manipulation,'' in \emph{Proceedings of IEEE/RSJ International Conference on Intelligent Robots and Systems. IROS'96}, vol.~2.\hskip 1em plus 0.5em minus 0.4em\relax IEEE, 1996, pp. 889--896.

\bibitem{lynch1999dynamic}
------, ``Dynamic nonprehensile manipulation: Controllability, planning, and experiments,'' \emph{The International Journal of Robotics Research}, vol.~18, no.~1, pp. 64--92, 1999.

\bibitem{srinivasa2005using}
S.~S. Srinivasa, M.~A. Erdmann, and M.~T. Mason, ``Using projected dynamics to plan dynamic contact manipulation,'' in \emph{2005 IEEE/RSJ International Conference on Intelligent Robots and Systems}.\hskip 1em plus 0.5em minus 0.4em\relax IEEE, 2005, pp. 3618--3623.

\bibitem{ichnowski2022gompfit}
J.~Ichnowski, Y.~Avigal, Y.~Liu, and K.~Goldberg, ``Gomp-fit: Grasp-optimized motion planning for fast inertial transport,'' in \emph{2022 International Conference on Robotics and Automation (ICRA)}, 2022, pp. 5255--5261.

\bibitem{pham2018fastsuction}
\BIBentryALTinterwordspacing
H.~Pham and Q.~Pham, ``Critically fast pick-and-place with suction cups,'' \emph{CoRR}, vol. abs/1809.03151, 2018. [Online]. Available: \url{http://arxiv.org/abs/1809.03151}
\BIBentrySTDinterwordspacing

\bibitem{zeng2020tossingbot}
A.~Zeng, S.~Song, J.~Lee, A.~Rodriguez, and T.~Funkhouser, ``Tossingbot: Learning to throw arbitrary objects with residual physics,'' \emph{IEEE Transactions on Robotics}, vol.~36, no.~4, pp. 1307--1319, 2020.

\bibitem{wang2020swingbot}
C.~Wang, S.~Wang, B.~Romero, F.~Veiga, and E.~Adelson, ``{SwingBot: Learning Physical Features from In-hand Tactile Exploration for Dynamic Swing-up Manipulation},'' 2020.

\bibitem{zhang2021rotla}
H.~Zhang, J.~Ichnowski, D.~Seita, J.~Wang, and K.~Goldberg, ``{Robots of the Lost Arc: Learning to Dynamically Manipulate Fixed-Endpoint Ropes and Cables},'' 2021.

\bibitem{chi2022iterativeresidualpolicygoalconditioned}
\BIBentryALTinterwordspacing
C.~Chi, B.~Burchfiel, E.~Cousineau, S.~Feng, and S.~Song, ``Iterative residual policy: for goal-conditioned dynamic manipulation of deformable objects,'' 2022. [Online]. Available: \url{https://arxiv.org/abs/2203.00663}
\BIBentrySTDinterwordspacing

\bibitem{curobo_report23}
B.~Sundaralingam, S.~K.~S. Hari, A.~Fishman, C.~Garrett, K.~V. Wyk, V.~Blukis, A.~Millane, H.~Oleynikova, A.~Handa, F.~Ramos, N.~Ratliff, and D.~Fox, ``curobo: Parallelized collision-free minimum-jerk robot motion generation,'' 2023.

\bibitem{luh1980line}
J.~Y. Luh, M.~W. Walker, and R.~P. Paul, ``On-line computational scheme for mechanical manipulators,'' 1980.

\bibitem{berscheid2021jerk}
L.~Berscheid and T.~Kr{\"o}ger, ``Jerk-limited real-time trajectory generation with arbitrary target states,'' \emph{Robotics: Science and Systems XVII}, 2021.

\bibitem{mcfee2015librosa}
B.~McFee, C.~Raffel, D.~Liang, D.~P. Ellis, M.~McVicar, E.~Battenberg, and O.~Nieto, ``librosa: Audio and music signal analysis in python,'' in \emph{Proceedings of the 14th python in science conference}, 2015, pp. 18--25.

\bibitem{2020SciPy-NMeth}
P.~Virtanen, R.~Gommers, T.~E. Oliphant, M.~Haberland, T.~Reddy, D.~Cournapeau, E.~Burovski, P.~Peterson, W.~Weckesser, J.~Bright, S.~J. {van der Walt}, M.~Brett, J.~Wilson, K.~J. Millman, N.~Mayorov, A.~R.~J. Nelson, E.~Jones, R.~Kern, E.~Larson, C.~J. Carey, {\.I}.~Polat, Y.~Feng, E.~W. Moore, J.~{VanderPlas}, D.~Laxalde, J.~Perktold, R.~Cimrman, I.~Henriksen, E.~A. Quintero, C.~R. Harris, A.~M. Archibald, A.~H. Ribeiro, F.~Pedregosa, P.~{van Mulbregt}, and {SciPy 1.0 Contributors}, ``{{SciPy} 1.0: Fundamental Algorithms for Scientific Computing in Python},'' \emph{Nature Methods}, vol.~17, pp. 261--272, 2020.

\bibitem{tim_sainburg_2019_3243139}
\BIBentryALTinterwordspacing
T.~Sainburg, ``timsainb/noisereduce: v1.0,'' Jun. 2019. [Online]. Available: \url{https://doi.org/10.5281/zenodo.3243139}
\BIBentrySTDinterwordspacing

\bibitem{sainburg2020finding}
T.~Sainburg, M.~Thielk, and T.~Q. Gentner, ``Finding, visualizing, and quantifying latent structure across diverse animal vocal repertoires,'' \emph{PLoS computational biology}, vol.~16, no.~10, p. e1008228, 2020.

\bibitem{kirillov2023segany}
A.~Kirillov, E.~Mintun, N.~Ravi, H.~Mao, C.~Rolland, L.~Gustafson, T.~Xiao, S.~Whitehead, A.~C. Berg, W.-Y. Lo, P.~Doll{\'a}r, and R.~Girshick, ``Segment anything,'' \emph{arXiv:2304.02643}, 2023.

\end{thebibliography}
\vspace{12pt}

\end{document}